\documentclass[10pt,twocolumn,letterpaper]{article}

\usepackage{cvpr}
\usepackage{times}
\usepackage{epsfig}
\usepackage{graphicx}
\usepackage{amsmath}
\usepackage{amssymb}
\usepackage{multirow}
\usepackage{authblk}

\usepackage[breaklinks=true,bookmarks=false]{hyperref}

\cvprfinalcopy 

\def\cvprPaperID{1271} 

\ifcvprfinal\fi
\begin{document}

\title{Dual Skipping Networks}
\newcommand*\samethanks[1][\value{footnote}]{\footnotemark[#1]}
\author{Changmao Cheng$^{1}$\thanks{Equal contribution. Email: \{cmcheng16,yanweifu\}@fudan.edu.cn}, Yanwei Fu$^2$\samethanks, Yu-Gang Jiang$^1$\thanks{Corresponding author. Email: ygj@fudan.edu.cn},\vspace{-0.13in}\\
 Wei Liu$^3$, Wenlian Lu$^4$, Jianfeng Feng$^4$, Xiangyang Xue$^1$
 
\small{$^1$Shanghai Key Lab of Intelligent Information Processing, School of Computer Science, Fudan University}\\
$^2$ School of Data Science, Fudan University\qquad$^3$ Tencent AI Lab\qquad$^4$ ISTBI, Fudan University\\
}


\maketitle
\thispagestyle{empty}

\begin{abstract}
Inspired by the recent neuroscience studies on the left-right asymmetry of the human brain in processing low and high spatial frequency information, this paper introduces a dual skipping network which carries out coarse-to-fine object categorization. Such a network has two branches to simultaneously deal with both coarse and fine-grained classification tasks. Specifically, we propose a layer-skipping mechanism that learns a gating network to predict which layers to skip in the testing stage. This layer-skipping mechanism endows the network with good flexibility and capability in practice. Evaluations are conducted on several widely used coarse-to-fine object categorization benchmarks, and promising results are achieved by our proposed network model.
\end{abstract}

\section{Introduction}
\begin{figure}
	\centering
	\small
	\begin{tabular}{c}
		\includegraphics[scale=0.12]{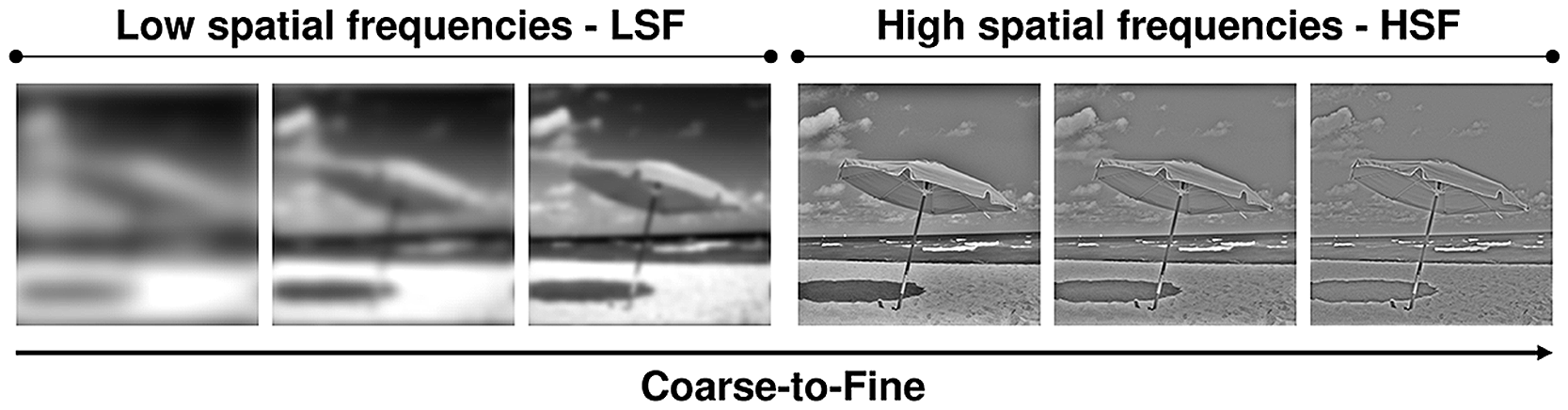}\tabularnewline
		(1)\tabularnewline
		\includegraphics[scale=0.23]{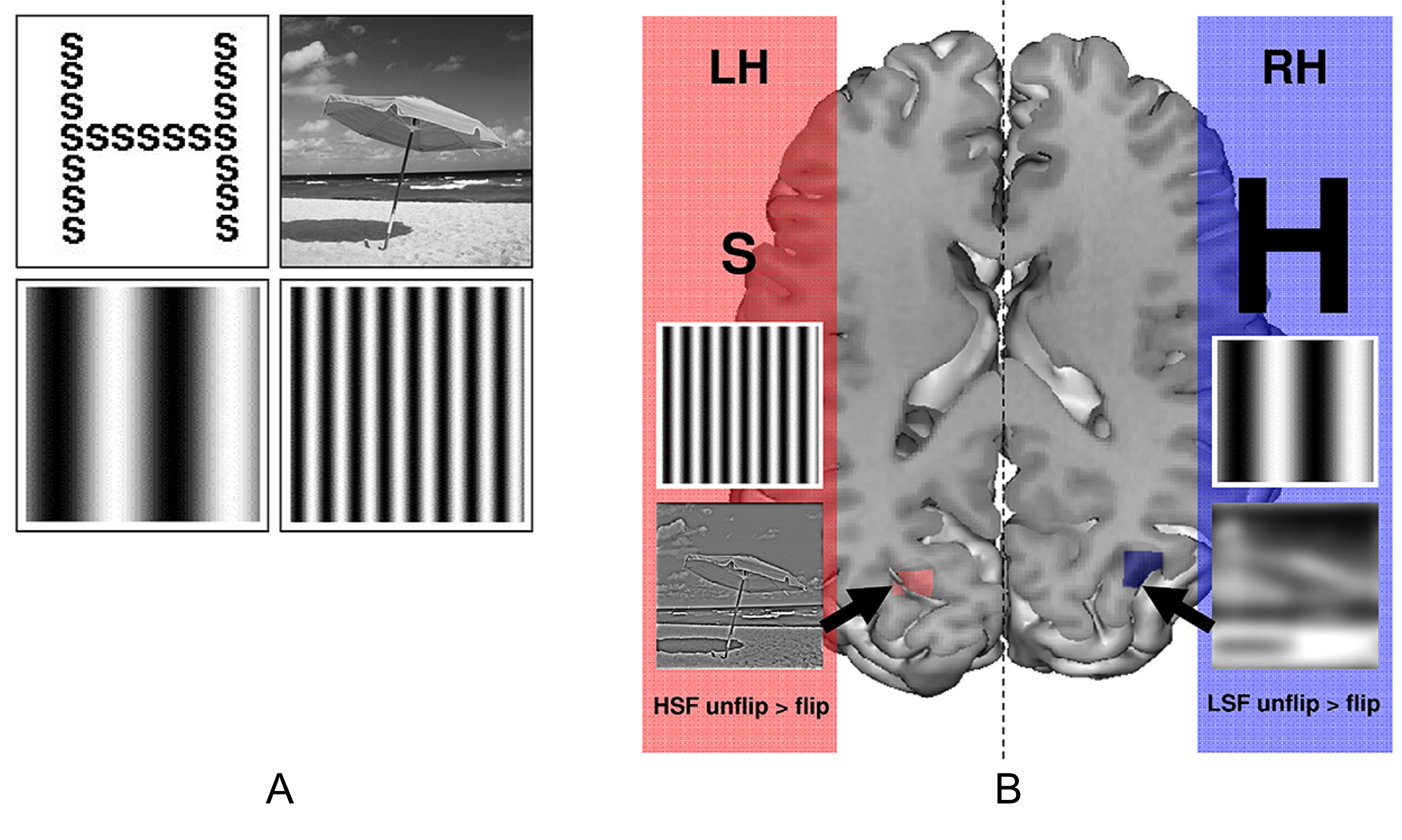}\tabularnewline
		(2)\tabularnewline
	\end{tabular}\caption{\label{fig:overview}(1) The coarse-to-fine sequence. (2-A) Example images to assess cerebral asymmetries
		for spatial frequencies. (2-B) Hemispheric specialization: the left
		hemisphere (LH)/the right hemisphere (RH) are predominantly involved
		in the local/global letter identification. Figures from \cite{kauffmann2014neural}.}
\end{figure}
Though there are still lots of arguments towards the exactly
where and how visual analysis is processed within the human brain, there
is considerable evidence showing that visual analysis takes place
in a predominately and default coarse-to-fine sequence \cite{kauffmann2014neural}
as shown in Fig. \ref{fig:overview}(1). The coarse-to-fine
perception is also proportional to the length of the cerebral circuit
path, i.e. time. For example, when the image
of Fig. \ref{fig:overview}(1) is very quickly shown to a person, only very coarse visual stimuli can be perceived, such as sand and umbrella, which
is usually of low spatial frequencies. Nevertheless, given
a longer duration, fine-grained details with relatively higher spatial frequencies
can be perceived. It is natural to ask whether our network has such
a mechanism of predicting coarse information with short paths, and
fine visual stimuli with longer paths.

Another question is how the coarse-to-fine sequence is processed
in the human brain? Recent biological experiments \cite{kauffmann2014neural,asymmetrical_NMDA,left_right_hippocampal} reveal that functions
of the two cerebral hemispheres are not exactly the same in processing of spatial frequency information. The
left hemisphere (LH) and the right hemisphere (RH) are predominantly involved
in the high and low spatial frequency processing respectively. As illustrated
in Fig. \ref{fig:overview}(2-A), the top-left figure is a large global
letter made up of small local letters, called a Navon figure; the top-right figure is a scene
image. The bottom-left and bottom-right are sinusoidal
gratings for the images above. In Fig.~\ref{fig:overview}(2-B),
given the same input visual stimuli, the highly activated regions in
LH and RH are corresponding to high (in red) and low (in blue) spatial
frequencies. Additionally, from the view of the evolution process, the left-right
asymmetry of the brain structure may be mostly caused by the long-term asymmetrical
functions performed~\cite{left_right_origin}. 

\begin{figure}[t]
	\centering{}\includegraphics[width=0.5\textwidth]{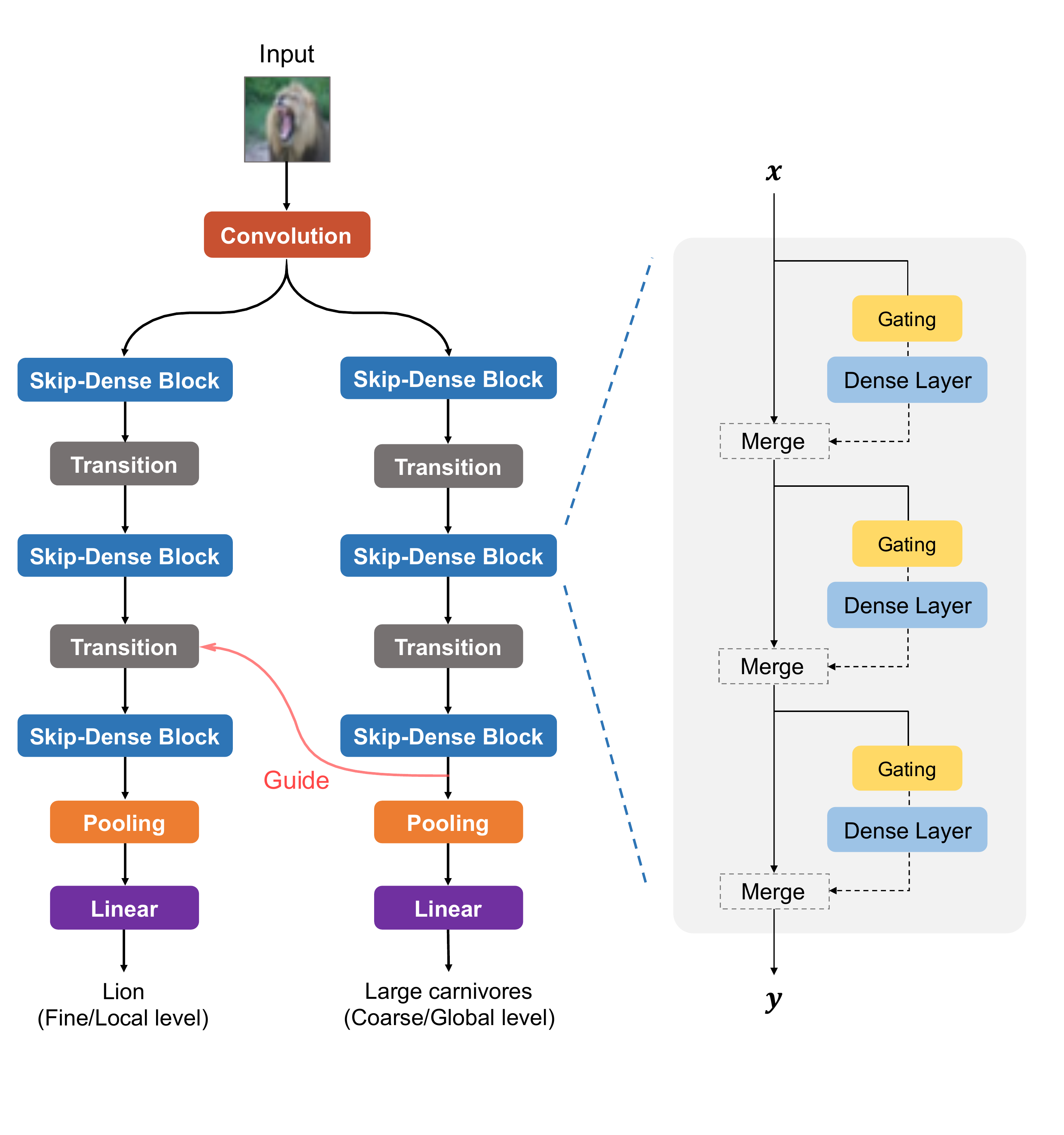}
	{\label{fig:model}\caption{\textbf{Dual Skipping Networks: the left-right asymmetric architecture.} (i) Two homogeneous subnets derived from DenseNets process input visual
			information asymmetrically. One branch is to predict coarse/global
			level object classes, the other is to predict fine/local level object
			classes. They share the same preliminary visual processing module,
			namely the shared convolutional layer. Each subnet is stacked mainly
			by abstraction blocks and transition layers iteratively. Linear layers transform features into predictions. The top-down guide link delivers feedback information from a high abstraction level of the coarse/global subnet to a lower abstraction level of the fine/local one. (ii) Each skip-dense block contains several skippable
			densely connected convolutional layers controlled by a cheap affiliated
			gating network.}
	}
\end{figure}

Some researchers \cite{hassabis2017neuroscience} believe that biological
plausibility can be used as a guide to design intelligent systems. In the light of this understanding and to mimic the hemispheric specialization,
we propose a dual skipping network, which is a
left-right asymmetric layer skippable
network. Our network can enable the coarse-to-fine object categorization
in a single framework. The whole network is structured in Fig.~\ref{fig:model}.
Our network has two branches by referring to LH and RH respectively.
Both branches have roughly the same initialized layers and structures.
The networks are built by stacking skip-dense blocks, namely groups of densely connected convolutional layers which can be dropped dynamically. The unique connections are built by learning
varying knowledge or abstraction. Transition layers aim at manipulating the capacity of features learned from preceding layers. The functionality of each branch
is ``memorized'' from the given input and supervised information
in the learning stage. The ``Guide'' arrow refers to a top-down facilitation of recognition that feeds the high-level information from the coarse branch to relatively lower-level visual processing modules of the fine branch inspired by a similar mechanism in the brain \cite{kauffmann2014neural,peyrin_fmri,integrated_v1}. Though spatial frequency cannot be equated with the granularity of recognition, the dual skipping network might work similar to hemispheric specialization depending on the granularity of supervised information.

The proposed \emph{layer-skipping mechanism}
for a single input is to utilize only a part of layers in the deep model
for the purpose of computation sparsity and flexibility. The organisms
like humans tend to use their energy ``wisely'' for the recognition and categorization task given
visual stimuli \cite{left_right_origin}. Some recent studies \cite{mapping_brain}
in neuroscience also showed that the synaptic cross-layer connectivity
is common in the human neural system, especially in the same abstraction
level. In contrast, classical deep convolutional neural
networks (e.g. AlexNet \cite{deng2009imagenet}, VGG \cite{simonyan2014very}) do not have this
mechanism and have to run the entire network at inference time. On the other hand,
the recent study \cite{Bolukbasi2017AdaptiveNN} found that most
of data samples are easy to be correctly classified without the utilization
of very deep networks. In particular, we propose
an affiliated gating network that learns to predict whether skipping several convolutional
layers in the testing stage. Our networks are evaluated on three datasets
in the coarse-to-fine object recognition tasks. The results show the
effectiveness of the proposed network.

\noindent \textbf{Contributions.}\quad
In this paper, inspired by
the left-right asymmetry of the brain, we propose a dual skipping network. The novelties come from several points: (1)
The left and right branch network structures towards solving coarse-to-fine
classification are proposed. Our network is inspired by the
recent theory in neuroscience \cite{kauffmann2014neural}. (2) A novel
\emph{layer-skipping mechanism} is introduced to
skip some layers at the testing stage. (3) We employ the top-down feedback facilitation to guide the fine-grained classification via high-level global semantic information. (4) Additionally, we create a novel dataset named small-big
MNIST (sb-MNIST) dataset with the hope of facilitating the research
on this topic. 
\section{Related Work}

\noindent \textbf{Hemispheric Specialization.}\quad{}Left-right asymmetry
of the brain has been widely studied through psychological examination
and functional imaging in primates \cite{mapping_brain}. Recent research
showed that both shapes of neurons and density of neurotransmitter
receptor expression depend on the laterality of presynaptic origin
\cite{asymmetrical_NMDA,left_right_hippocampal}. There is also a
proof in terms of information transfer that indicates the connectivity
changes between and within the left and right inferotemporal cortexes
as a result of recognition learning \cite{granger_face_recognition}.
Learning also differs in both local and population as well as theta-nested
gamma frequency oscillations in both hemispheres and there is greater
synchronization of theta across electrodes in the right IT than the
left IT \cite{theta_cortex}. It has recently been argued that the
left hemisphere specializes in controlling routine and tends to focus
on local aspects of the stimulus while the right hemisphere specializes
in responding to unexpected stimuli and tends to deal with the global
environment \cite{Turgeon1994}. For more information, we refer to
one recent survey paper \cite{kauffmann2014neural}. 

\noindent \textbf{Deep Architectures.}\quad
Starting with the notable victory of AlexNet \cite{KrizhevskySH12}, ImageNet \cite{deng2009imagenet}
classification contest has boomed the exploration of deep CNN architectures.
Later, \cite{He2016DeepRL} proposed deep Residual Networks (ResNets)
which mapped lower-layer features into deeper layers by shortcut connections
with element-wise addition, making training up to hundreds or even
thousands of layers feasible. Prior to this work, Highway Networks
\cite{srivastava2015highway} devised shortcut connections with input-dependent
gating units. Recently, \cite{Huang2016DenselyCC} proposed a compact
architecture called DenseNet that further integrated shortcut connections
to make early layers concatenated to later layers. The simple dense
connectivity pattern surprisedly achieves the state-of-the-art accuracy
with fewer parameters. Different from the shortcut connections used
in DenseNets and ResNets, our layer-skipping mechanism is learned to
predict whether skipping one particular layer in the testing stage.
The skipping layer mechanism is inspired by the fact that on one-time
cognition process, only $1\%$ of total neurons in the human brain
are used \cite{left_right_origin}.

\noindent \textbf{Coarse-to-Fine Recognition.}\quad
The coarse-to-fine recognition process is natural and favored by researchers \cite{yan2015hd, Zamir2016FeedbackN, mottaghi2015coarse, cheng2015learning} and it is very useful in real-world applications. Feedback Networks \cite{Zamir2016FeedbackN} developed a coarse-to-fine representation via recurrent convolutional
operations; such that the current iteration's output gives a feedback
to the prediction at the next iteration. With the different emphasis
on biological mechanisms, our design is also a coarse-to-fine formulation
but in a feedforward fashion. Additionally, our coarse-level categorization
will guide the fine-level task. HD-CNN \cite{yan2015hd} learned a hierarchical structure of classes by grouping fine categories into coarse classes and embedded deep CNNs into the category hierarchy for better fine-grained prediction. In contrast, our two-branch model deals with coarse 
and fine-grained classes simultaneously. Relatedly, research on fine-grained classification \cite{fu2017look,xiao2015application,lin2015bilinear,zhang2016picking,wang2015multiple} has been drawing a lot of attention over the years.

\noindent \textbf{Conditional Computation.}\quad
Some efforts have been made to bypass the computational cost of deep models in the testing
stage, such as network compression \cite{Zhang2017ShuffleNetAE} and conditional computation (CC) \cite{Bengio2013EstimatingOP}. The
CC often refers to the input-dependent activation for neurons or unit
blocks, resulting in partial involvement fashion for neural networks.
The CC learns to drop some data points or blocks within a feature
map and thus it can be taken as an adaptive variant of dropout. \cite{Bengio2013EstimatingOP}
introduced Stochastic Times Smooth neurons as binary gates in
a deep neural network and termed the straight-through estimator whose
gradient is learned by heuristically back-propagating through the
threshold function. \cite{Ba2013AdaptiveDF} proposed a `standout'
technique which uses an auxiliary binary belief network to compute
the dropout probability for each node. \cite{Bengio2015ConditionalCI}
tackled the problem of selectively activating blocks of units via
reinforcement learning. Later, \cite{Odena2017ChangingMB} used a
Recurrent Neural
Network (RNN) controller to examine and constrain intermediate activations of
a network at test-time. \cite{Figurnov2016SpatiallyAC} incorporates
attention into ResNets for learning an image-dependent early-stop policy
in residual units, both in layer level and feature block level. Once stopped, the following layers in the same layer group will not be executed. While our mechanism is to predict whether each one particular layer should be skipped or not, it assigns more selectivity for the forward path.

Compared with conditional computation, our layer-skipping mechanism
is different in two points: (1) CC learns to drop out some units in feature maps, whilst our gating network learns to
predict whether skipping the layers. (2) CC usually
employs reinforcement learning algorithms which have in-differentiable
loss functions and need huge computational cost for policy search; in contrast, our gating network is a differentiable function
which can be used for individual layers in our network.

\section{Model}

The presented dual skipping network is overviewed in Fig. \ref{fig:model}. The two subsets are shared with the same visual inputs and built upon several types of modules,
namely, the shared convolutional layer, skip-dense blocks, transition layers, pooling and classification
layers. The motivation and structure of each building block are discussed in this section.

\noindent \textbf{Shared convolutional layer.}\quad
The input image is
firstly processed by the convolutional layer (shared by two following
subnets) to extract the low-level visual signal. Such convolutional
layers can be biologically corresponding to the primary visual cortex
V1 \cite{visual_v1}. The two left and right subnets solve the
fine-grained and coarse classification respectively.

\noindent \textbf{Left/right subnet.}\quad
Each subnet is stacked mainly
by skip-dense blocks and transition layers iteratively. As in Fig.~\ref{fig:model}~(right), we define the skip-dense block which can
be viewed as a level of visual concept abstraction; and the gating
network is learned whether to block the information flow passing to
dense layer. Each transition has convolutional
operations with $1\times1$ filter size and pooling, which aims at changing the
number and spatial size of feature maps for the next skip-dense
block. Both the left and right subnets are almost equivalent in the
general structure.

\noindent \textbf{Pooling and classification layers.} The global average pooling
and linear classifier are added at the top of each branch
for the prediction tasks.

\subsection{Skip-Dense Block}
We give the structure details of a skip-dense block in Fig.~\ref{fig:model}~(right). For the input pattern $\mathbf{x}$ from
a shallower abstraction level, a shortcut connection is applied after
each dense layer. And the option of information flowing through each dense layer
is checked by a cheap gating network which is input-dependent. The output pattern $\mathbf{y}$ is then processed into the next transition
layer and entered into the next abstraction level.

\noindent \textbf{Dense Layer.} \quad This type of convolutional layers
comes from DenseNet \cite{Huang2016DenselyCC} or ResNet \cite{He2016DeepRL}. The difference between DenseNet and ResNet comes from the different combination methods of shortcut connections, namely merge functions here. Therefore, the merge operation of preceding layers can be \emph{channel concatenation} or \emph{element-wise
	addition}, used in DenseNets \cite{Huang2016DenselyCC}
and ResNets \cite{He2016DeepRL} respectively. In particular, the pre-activation
unit \cite{He2016IdentityMI} facilitates both of the merge functions
from our observation, thus used in our proposed model. In general, the dense connectivity
pattern enables any layer to more easily access proceeding layers and thus
use make individual layers additionally supervised from the shorter
connections. 

Residual networks have been found behaving like ensembling numerous shallower paths \cite{Veit2016ResidualNB}. Idling few layers in these networks may not dramatically
degrade the performance. Intuitively, the tremendous number of
hidden paths may be redundant; an efficient path selection mechanism
is essential if we want to reduce the computational cost in the testing
stage. Mathematically, there are exponentially many hidden
paths in these networks. For example, for $L$ dense layers, we can
obtain $2^{L}$ hidden paths. Dynamic path selection could pursue the specialized and flexible formations of nested convolution operations. In neuroscience, the experimental evidence \cite{integrated_v1,peyrin_fmri}
also indicates that the path optimization may exist in the parvocellular
pathways of the left and right hemispheres when the low and high-pass
signals are processed. This supports the motivation of our path selection
from the viewpoint of neuroscience.

\noindent \textbf{Gating network.}\quad
The gating network is introduced
for path selection. Particularly, our gating network is learned
to judge whether or not skipping the convolutional layer from the training
data. It can also be taken as one special type of regularizations:
the gating network should be
inclined not to skip too many layers if the input data is complex and vice versa. Here, we utilize
a $N\times1$ fully-connected layer for the $N$-dimensional input
features and then a threshold function is applied to the scalar output.
We preprocess the input features by average pooling in practice. The
parameters of the fully-connected layer are learned from the training
set and the threshold function is designed to control the learning
process. 

\noindent \textbf{Threshold function of gating network}. The policy
of designing and training the threshold function as an estimator is
very critical to the success of the skipping mechanism. Luckily, the magnitude
of a unit often determines its importance for the categorization task
in CNNs. Based on this observation, we derive a simple end-to-end
training scheme for the entire network. Specifically, the output of
threshold function is multiplied with each unit of the convolutional
layer output, which affects the layer importance of categorization.

The key ingredient of gating network is the threshold function.
Given an activation or input, the threshold function needs to judge
whether skipping the following dense layer or not as in Fig.~\ref{fig:model}
~(right). Intuitively, the threshold function performs as a binary
classifier. In here, we choose hard sigmoid function
\begin{equation}
\mathsf{hard\ sigm}(x)=\max\left(0,\min\left(kx+\frac{1}{2},1\right)\right)\label{eq:hard_sigmoid}
\end{equation}
for its first derivative in $(0, 1)$ keeps constant, which encourages more flexible path searching compared with the sigmoid function. Besides, for the outputs clipped to 0 or 1, we borrow the idea of straight-through
estimator \cite{Bengio2013EstimatingOP} to make the error backpropagated
through the threshold function. Thus it is always differentiable. 

The slope variable $k$ is the key parameter to determine the output scaling of dense layers.
The $k$ is initialized at 1 and increased by a fixed value every
epoch. As results, the learned curve
of gating will be slope enough to make the outputs of gating network either 0 or 1 in the training process. It can be used as an approximation
of step function that emits binary decisions. However, the large slope variable would make the weight training of gating modules unstable. Again, we use the straight-through
estimator to keep $k$ equal to one in backward mode. 

The mutual adjustment and regularization of gating and dense layers
abbreviate the problem of training difficulty and hard convergence
of reinforcement learning \cite{Bengio2015ConditionalCI}. 
For inference, the gating network makes discrete binary decisions to save computation.

\subsection{Guide}
The faster coarse/global
subnet can guide the slower fine/local subnet with global context
information of objects in a top-down fashion, inspired from the LSF-based top-down facilitation of recognition in the visual cortex proposed by \cite{bar2006top}. In here, we select
the output features of the last skip-dense block in the coarse branch
to guide the last transition layer in the fine branch. Specifically,
the output features are bilinear upsampled and concatenated into the input features of the last
transition layer in the local subnet. The injection of feedback information
from the coarse level can be beneficial for the fine-grained object categorization.

\section{Experiments}

\noindent \textbf{Datasets.}\quad
We conduct experiments on four
datasets, namely sb-MNIST, CIFAR-100~\cite{Krizhevsky2009LearningML}
and CUB-200-2011 \cite{WahCUB_200_2011}, Stanford Cars~\cite{Krause20133DOR}.
\begin{figure}[t]
	\centering
	\small
	\begin{tabular}{cccc}
		\includegraphics[scale=0.2]{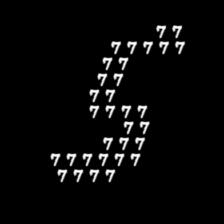}  & \includegraphics[scale=0.2]{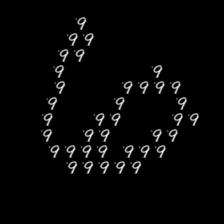}  & \includegraphics[scale=0.2]{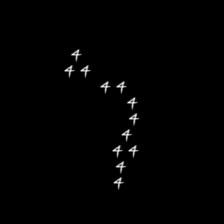}  & \includegraphics[scale=0.2]{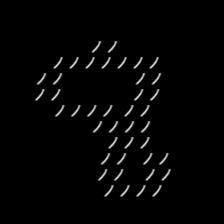}\tabularnewline
	\end{tabular}\caption{\label{fig:sb-MNIST}Illustrated examples of sb-MNIST dataset. Each
		``big'' figure is composed of copies of a ``small'' figure.}
\end{figure}

\noindent \textbf{sb-MNIST}. Inspired by the experiments with Navon figures in \cite{kauffmann2014neural},
we build sb-MNIST dataset by randomly selecting two images from
MNIST dataset and using the first one as the local figure to construct
the second one. We generate 120,000 training images and 20,000 testing
images for building the dataset\footnote{The CVPR version gives incorrect numbers due to typo.}. Some examples are illustrated
in Fig.~\ref{fig:sb-MNIST}. 

\noindent \textbf{CIFAR-100} has 60,000 images from 100 fine-grained
classes, which are further divided into 20 coarse-level classes. The
image size is $32\times32$. We use the standard training/testing
split\footnote{\url{https://www.cs.toronto.edu/~kriz/cifar.html}}. 

\noindent \textbf{CUB-200-2011} contains 11,788 bird images of 200
fine-grained classes. Strictly following the biological taxonomy,
we collect 39 coarse labels in total by the family names of the 200 bird
species. For instance, the black-footed albatrosses belong to Diomedeidae
family. We use the default training/testing split.

\noindent \textbf{Stanford Cars} is another fine-grained classification
dataset. It contains 16,185 car images of 196 fine classes (e.g. Tesla
Model S Sedan 2012 or Audi S5 Coupe 2012) which describe the properties
such as Maker, Model, Year of the car. In terms of the basic car types
defined in \cite{Zamir2016FeedbackN}, it can be categorized into
7 coarse classes containing Sedan, SUV, Coupe, Convertible, Pickup,
Hatchback and Wagon. The default training/test split is used here. 

\vspace{0.04in}

\noindent \textbf{Competitors.}\quad{}We compare the following several
baselines. (1) \emph{Feedback Net} \cite{Zamir2016FeedbackN}: as the
only work of enabling the coarse-to-fine classification, it is a feedback
based learning architecture in which each representation is formed
in an iterative manner based on the feedback received from previous iteration's
output. The network is instantiated using existing RNNs. Thus Feedback Net has a better classification performance
than the standard CNNs. (2)\emph{ DenseNet} \cite{Huang2016DenselyCC}:
DenseNet connects each layer to all of its preceding layers in a feed-forward
fashion; and the design of our skip-dense block is derived from DenseNet.
Thus comparing with DenseNet, our network has two new components:
the gating network to skip some layers dynamically and the two-branch
structure for solving the coarse-to-fine classification. (3)\emph{ ResNet}
\cite{He2016DeepRL}: it is an extension of traditional CNNs by learning
the residual of each layer to enable the network of being trained
substantially deeper than previous CNNs. 

\vspace{0.02in}

\noindent \textbf{Implementation details.}\quad
The merge types of our models can be \emph{channel concatenation} or \emph{element-wise addition}, ``Concat'' and ``Add'' for short.
In the experiments, we configure our model based on Concat merge type, namely DenseNets as default. The shared convolution layer with
output channels of twice the growth rate is performed to the input
visual images. We replace the outputs of skipped dense layers with features maps of the same size filled with zero at inference.

For sb-MNIST, we configure 4 skip-dense blocks each with 3 dense layers and growth rate as 6. We do not use guide link for sb-MNIST dataset.
For CIFAR-100, we verify our method with different model settings. Two settings of Concat merge type are based on DenseNet-40 and DenseNet-BC-100 \cite{Huang2016DenselyCC}, denoted as Concat-40 and Concat-BC-100 respectively. We also report the results of our model with Add merge type. Referred as Add-166, the model is built very similar to ResNet-164 \cite{He2016IdentityMI}, except for Add-166 including two extra transition layers. Data preprocessing procedure and initializations follow \cite{Huang2016DenselyCC,He2016IdentityMI}. 

For CUB-200-2011 and Stanford Cars, our model is built on DenseNet-121 \cite{Huang2016DenselyCC}. The input images are resized to $360\times360$ for training and test on both datasets. We incorporate the ImageNet pre-trained DenseNet-121 weights into our model as network initialization. Mini-batch size is set to 16, learning rate is started with 0.01. To save GPU memory, we use the memory-efficient implementation of DenseNets~\cite{Pleiss2017MemoryEfficientIO} here. The experiments are conducted without using any bounding
boxes or part annotations.

All of our models are trained using SGD with a cosine annealing learning rate schedule \cite{loshchilov2016sgdr} and Nesterov momentum \cite{Sutskever2013OnTI} of 0.9 without dampening. The models are trained jointly with or without gating modules, then fine-tuned with gating. We found that joint training as a start speeds up the whole process and makes the performance more stable. We do gradient clipping with $L_2$ norm threshold 1.0 for avoiding gradient explosion. For the gating modules whose outputs are all 0 or 1 in a batch, we apply auxiliary binary cross entropy loss on them to guarantee functioning of the gating modules. All the code is implemented in PyTorch and run on Linux machines equipped with the NVIDIA GeForce GTX 1080Ti graphics cards.

\subsection{Main Results and Discussions}
\begin{table}
	\centering
	\small
	\begin{tabular}{c|cc}
		\hline 
		\multirow{2}{*}{Methods } & \multicolumn{2}{c}{{Accuracy ($\%$)}}\tabularnewline
		\cline{2-3} 
		& {Local}  & {Global}\tabularnewline
		\hline 
		\hline
		{LeNet \cite{LeCun1998GradientbasedLA}}  & 90.3  & 70.2\tabularnewline
		\cline{1-1} 
		{DenseNet }  & 98.3  & 97.2\tabularnewline
		\cline{1-1} 
		{ResNet }  & 98.1  & 96.8\tabularnewline
		\cline{1-1} 
		{Ours }  & \textbf{99.1}  & \textbf{99.1}\tabularnewline
		\hline 
	\end{tabular}{\caption{\label{tab:Results-of-sb-MNIST}Results on sb-MNIST dataset. Our network
			has 17 layers and averagely {\raise.17ex\hbox{${\scriptstyle \sim}$}}$20\%$
			layers can be skipped in the testing step. DenseNet and ResNet have
			40 and 18 layers respectively.}
	}
\end{table}
\subsubsection{sb-MNIST }

The results on this dataset are compared in Tab. \ref{tab:Results-of-sb-MNIST}.
The ``Local'' and ``Global'' indicate the classification of ``small''
figures and ``big'' figures. On this dataset, DenseNet, ResNet and
LeNet are run separately for the both ``Local'' and ``Global''
tasks. The classification of ``small'' and ``big'' figures in the
image corresponds to the identification task of high and low spatial
frequency processing \cite{kauffmann2014neural}. This toy dataset gives us a general understanding of it.

Judging from the results in Tab. \ref{tab:Results-of-sb-MNIST}, our
method outperforms other methods by a clear margin.
It is largely due to two reasons. Firstly, the skip-dense blocks can
efficiently learn the visual concepts; secondly, the gating network learns to find optimal path routing for avoiding suffering from both overfitting and underfitting. The performance of original version LeNet is
greatly suffered from the small number of layers and convolutional filters while DenseNet and ResNet show slight overfitting. Averagely, our model uses 76\% of all the parameters as 15\% and 31\% of dense layers in ``Local" and ``Global'' branches are skipped.
Due to a portion of layers skipped in testing step, the total running
time and running cost are lower.

\begin{table}[t]
	\centering
	\small
	\begin{tabular}{c|cc|c}
		\hline 
		\multirow{2}{*}{Methods} & \multicolumn{2}{c|}{Accuracy ($\%$)} & \multirow{2}{*}{Params}\tabularnewline
		\cline{2-3} 
		& {Local}  & {Global} & \tabularnewline
		\hline 
		\hline 
		\cite{icml2015_bayesian}  & 72.6  & \textendash{} & \textendash{}\tabularnewline
		\cline{1-1} 
		All-CNN \cite{Springenberg2014StrivingFS}  & 66.3  & \textendash{} & \textendash{}\tabularnewline
		\cline{1-1} 
		Net-in-Net \cite{netinnet} & 64.3 & \textendash{} & \textendash{}\tabularnewline
		\cline{1-1} 
		Deeply Sup. Net \cite{deepSupervised} & 65.4 & \textendash{} & \textendash{}\tabularnewline
		\cline{1-1} 
		FractalNet \cite{fractalnet} & 76.7 & \textendash{} & 38.6M\tabularnewline
		\cline{1-1} 
		Highway \cite{srivastava2015highway} & 67.6 & \textendash{} & \textendash{}\tabularnewline
		\cline{1-1} 
		{Feedback Net \cite{Zamir2016FeedbackN}}  & 71.1  & 80.8 & 1.9M \tabularnewline
		\cline{1-1} 
		{DenseNet-40 \cite{Huang2016DenselyCC}}  & 75.6  & 80.9 & 1.0M$\times$2\tabularnewline
		\cline{1-1} 
		{DenseNet-BC-100 \cite{Huang2016DenselyCC}}  & 77.7  & 83.0 & 0.8M$\times$2\tabularnewline
		\cline{1-1} 
		{ResNet-164 \cite{He2016IdentityMI}}  & 75.7  & 81.4 & 1.7M$\times$2\tabularnewline
		\cline{1-1}
		{Concat-40}  & \textbf{75.4}  & \textbf{82.9} & 0.9M+0.7M\tabularnewline
		\cline{1-1}
		{Concat-BC-100}  & \textbf{76.9}  & \textbf{83.4} & 0.7M+0.6M\tabularnewline
		\cline{1-1}
		{Add-166}  & \textbf{75.8}  & \textbf{82.5} & 1.5M+0.5M\tabularnewline
		\hline 
	\end{tabular}{\caption{\label{tab:Results-of-cifar-100}Results on CIFAR-100 dataset. The
			virtual depth of Feedback Net is 48, indicating the number of unfolded layers in Feedback
			Net, reported from \cite{Zamir2016FeedbackN}. ``$\times$2'' indicates two separate branches for two tasks. $a+b$ in the ``Params'' column represents the averagely used parameters $a$, $b$ in Local and Global branches of our models. The Global results of DenseNet and ResNet are run by ourselves.
	}}
\end{table}
\subsubsection{CIFAR-100}
We compare the results in Tab. \ref{tab:Results-of-cifar-100}. We
compare Feedback Net, DenseNet and ResNet as well as other previous
works. The ``Local'' and ``Global'' indicate the classification
tasks at fine-grained and coarse levels individually. DenseNet and
ResNet are run separately for the ``Local'' and ``Global''
tasks respectively.

In Tab. \ref{tab:Results-of-cifar-100}, our models
outperform several baselines and show comparable performances to the corresponding base models with much less computational
cost. Compared with the Feedback Net, Concat-40 gains the accuracy margins of $4.3\%$
and $2.1\%$ on ``Local'' and ``Global'' individually with relatively fewer parameters. This implies that our network has the better
capability of learning the coarse-to-fine information by the two branches. Though Feedback Net has a shallower physical depth 12, it contains more parameters and consumes more computation cost. This difference is largely caused by the
different architectures of two types of networks: Feedback Net
is built upon the recurrent neural network, while our network is a
forward network with the strategy of skipping some layers. Thus, our
network is more efficient in term of computational cost and running
time. 

On the ``Global'' task, our network can still
achieve a modest improvement over other baselines. One possible
explanation is that since the ``Global'' task is relatively easy,
the networks with standard layers may tend to overfit the training
data; in contrast, our gating network can skip some layers for better regularization.
\subsubsection{CUB-200-2011 and Stanford Cars datasets}

We compare the results on CUB-200-2011 and Stanford cars
datasets in Tab.~\ref{tab:Results-of-cub} and Tab.~\ref{tab:Results-of-stanford-car} respectively. The ``Local'' and ``Global'' tasks still refer to the fine-grained and coarse-level classification. Our network uses 121 layers and averagely {\raise.17ex\hbox{${\scriptstyle \sim}$}}$30\%$ and {\raise.17ex\hbox{${\scriptstyle \sim}$}}$13\%$ of dense
layers of global and local branches can be skipped in the testing step.

On the coarse classification, our result is better than those
baselines. On the fine-grained task, our models achieve results comparable to the state-of-the-art \cite{fu2017look} with much fewer parameters. On CUB-200-2011 dataset, our result is still better than that
of MG-CNN, Bilinear-CNN and RA-CNN (scale 1+2) which 
employ multiple VGG networks \cite{simonyan2014very} for classification, showing the effectiveness of our compact structure. Comparably, each
sub-network of Bilinear-CNN uses the pre-trained VGG network and
there is no information flow between two networks until the final
fusion layer.
\begin{table}
	\centering
	\small
	\begin{tabular}{c|cc|c}
		\hline 
		\multirow{2}{*}{Methods} & \multicolumn{2}{c|}{{Accuracy ($\%$)}}&\multirow{2}{*}{Params}\tabularnewline
		\cline{2-3} 
		& {Local}  & {Global}\tabularnewline
		\hline 
		\hline 
		{DenseNet-121 \cite{Huang2016DenselyCC}}  & 81.9 & 92.3 & 8M$\times$2\tabularnewline
		\cline{1-1} 
		{ResNet-34 \cite{He2016DeepRL}}  & 79.3 & 91.8 & 22M$\times$2\tabularnewline
		\cline{1-1}
		MG-CNN \cite{wang2015multiple} & 81.7 & \textendash{}&$\ge$144M\tabularnewline
		 \cline{1-1} 
		Bilinear-CNN \cite{lin2015bilinear} & 84.1 & \textendash{}&$\ge$144M\tabularnewline
		\cline{1-1} 
		RA-CNN (scale 1+2) \cite{fu2017look} & 84.7 & \textendash{}&$\ge$144M\tabularnewline
		\cline{1-1} 
		RA-CNN (scale 1+2+3) \cite{fu2017look} &\textbf{85.3}&\textendash{}&$\ge$144M\tabularnewline
		\hline 
		{Ours}  & 84.9 & \textbf{93.4} & 6M+5M\tabularnewline
		\hline 
	\end{tabular}{\caption{\label{tab:Results-of-cub}Results on CUB-200-2011 dataset. The results of DenseNet-121 and ResNet-34 are run by ourselves.
	}}
\end{table}
\begin{table}
	\centering
	\small
	\begin{tabular}{c|cc|c}
		\hline 
		\multirow{2}{*}{Methods} & \multicolumn{2}{c|}{Accuracy ($\%$)}&\multirow{2}{*}{Params}\tabularnewline
		\cline{2-3}
		& \multicolumn{1}{c|}{Local} & {Global}\tabularnewline
		\hline 
		\hline 
		{Feedback Net \cite{Zamir2016FeedbackN}}  & {53.4}  & 87.4 & 1.5M\tabularnewline
		\cline{1-1} 
		{DenseNet-121 \cite{Huang2016DenselyCC}}  & 90.5  &92.8 & 8M$\times$2\tabularnewline
		\cline{1-1} 
		{ResNet-34 \cite{He2016DeepRL}}  & 89.3  & 92.0 & 22M$\times$2\tabularnewline
		\cline{1-1} 
		Bilinear-CNN \cite{lin2015bilinear} & 91.3 &\textendash{}& $\ge$144M\tabularnewline
		\cline{1-1} 
		RA-CNN (scale 1+2) \cite{fu2017look} & {91.8} & \textendash{} & $\ge$144M\tabularnewline
		\cline{1-1} 
		RA-CNN (scale 1+2+3) \cite{fu2017look} & \textbf{92.5} & \textendash{} & $\ge$144M\tabularnewline
		\hline 
		{Ours }  & \multicolumn{1}{c|}{92.0 } & \textbf{93.8}& 6M+5M\tabularnewline
		\hline 
	\end{tabular}{\caption{\label{tab:Results-of-stanford-car}Results on Stanford Cars dataset. The results of DenseNet-121 and ResNet-34 are run by ourselves.}
	}
\end{table}

\subsection{Ablation Studies on CIFAR-100 dataset}

We also conduct some ablation studies to further evaluate and explain the
impacts on performances by choosing key settings in our models.

\subsubsection{Merge types and gating functions}

We compare the different choices of merge types and gating functions. For merge types, we use Concat-BC-100 and Add-166 for Concat and Add types respectively since the two models have roughly the same number of bottleneck dense layers (48 and 54). For gating functions, we compare the proposed $\mathsf{hard\ sigm}$ with $\mathsf{soft\ sigm}(x)=\frac{1}{1+e^{-x}}$.

The results are compared in Tab. \ref{tab:ablation-Results-on-Cifar-100}.
Two evaluation metrics have been used here, namely accuracy and skip
ratio. The skip ratio is computed by the number of skipped layers dividing
the total number of dense layers in each individual branch. 

Judging from the results in Tab. \ref{tab:ablation-Results-on-Cifar-100},
the combination of channel concatenation for merge and gating via
hard sigmoid is our best network configuration as it achieves the best performance with the fewest parameters. Besides, we notice
that $\mathsf{hard\ sigm}$ is generally better than $\mathsf{soft\ sigm}$ for both merge
types on most cases, indicating that the former possesses a stronger ability of routing path searching.

We compare the skip ratios on ``Local'' and ``Global'' tasks. In
testing stages, the coarse branch of our network will skip much more
layers than its sibling
fine branch. This is largely due to the fact that the ``Global'' task is easy,
and using relative a few layers of coarse branch is good enough to
grasp the coarse-level information. Interestingly, this point
follows the coarse-to-fine perception in the recent neural science
study \cite{kauffmann2014neural} that low spatial frequency information
is predominantly processed by right hemisphere faster with relatively shorter
paths in the cerebral system. 
\begin{table}
	\centering
	\small
	\begin{tabular}{c|c|cc|cc}
		\hline 
		\multirow{2}{*}{Merge} & \multirow{2}{*}{Gating} & \multicolumn{2}{c|}{Accuracy ($\%$)} & \multicolumn{2}{c}{Skip Ratio ($\%$)}\tabularnewline
		\cline{3-6} 
		&  & Local  & Global  & Local  & Global \tabularnewline
		\hline 
		\hline 
		\multirow{2}{*}{concat} & $\mathsf{hard}$  & 76.9  & 83.4  & 15.3  & 29.7 \tabularnewline
		\cline{2-6} 
		& $\mathsf{soft}$  & 73.8  & 79.7  & 15.4  & 30.5 \tabularnewline
		\hline 
		\multirow{2}{*}{add} & $\mathsf{hard}$  & 75.8  & 82.5  & 10.4  & 39.8 \tabularnewline
		\cline{2-6} 
		& $\mathsf{soft}$  & 74.2  & 83.1  & 6.1  & 23.5 \tabularnewline
		\hline 
	\end{tabular}\caption{\label{tab:ablation-Results-on-Cifar-100}Ablation study on CIFAR-100 dataset.
		`concat', `add',`hard',`soft' indicate the channel concatenation,
		element-wise addition, soft sigmoid function and hard sigmoid function.}
\end{table}

\begin{figure}[t]
	\centering
	\small
	\hbox{\hspace{2ex}\includegraphics[width=0.52\textwidth]{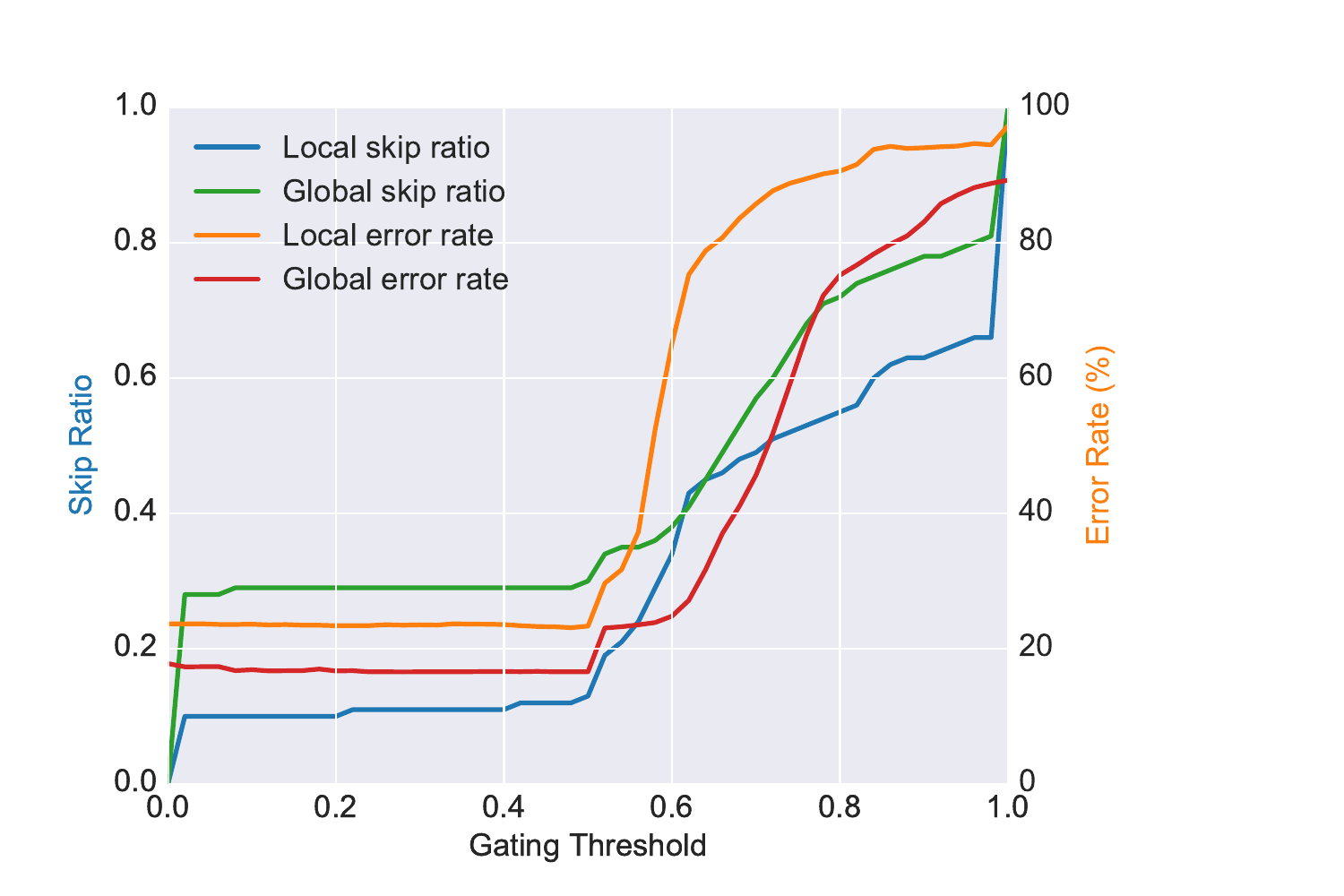}}
	\caption{Local and Global layer-skipping ratios and error rates under different gating thresholds. The slope variable is properly scaled for the smooth change of the skip ratios.}
	\label{fig:skip-ratio-acc} 
\end{figure}
\subsubsection{Skip ratio vs. error rate}
To further verify the flexibility of our model, we take the Concat-BC-100 model with $\mathsf{hard\ sigm}$ for gating as the study object since it achieves the excellent performance with very few parameters. We vary the threshold of gating
network which can affect the layer-skipping ratios and error rates of ``Local'' and ``Global''
tasks, as reported in Fig.~\ref{fig:skip-ratio-acc}.

From the results, we can conclude that both branches are capable of
producing acceptable accuracy within a certain range of gating thresholds. As expected,
the optimal range of skip ratios for two branches are not same  due to the different granularity level of recognition
tasks. For the Global branch, $0\%-35\%$ is the optimal skip ratio
range. For the Local branch, $0\%-20\%$ is the optimal skip ratio
range, which is more strict than the Global branch. The good thing
is that the learned gating module at each dense layer makes the performance
of two branches keep consistent under various user-defined thresholds. As the gating threshold value is raised above 0.5, the skip ratios of two branches increase rapidly and the error rates rise synchronously, meaning that the network becomes disordered and weak in expressivity. 
We also notice that the performances of two trained branches without
or with little layer skipping keep almost undamaged even we didn't train the full network without gating when fine-tuning. We conjecture two reasons behind this intriguing phenomenon. First, the shortcut connections make the model very stable and robust to the fine weight updating. The second reason is that there still exist some data samples using the whole dense layers when training with gating.

\subsection{Feature Visualization}
\begin{figure}
	\centering
	\includegraphics[width=0.46\textwidth]{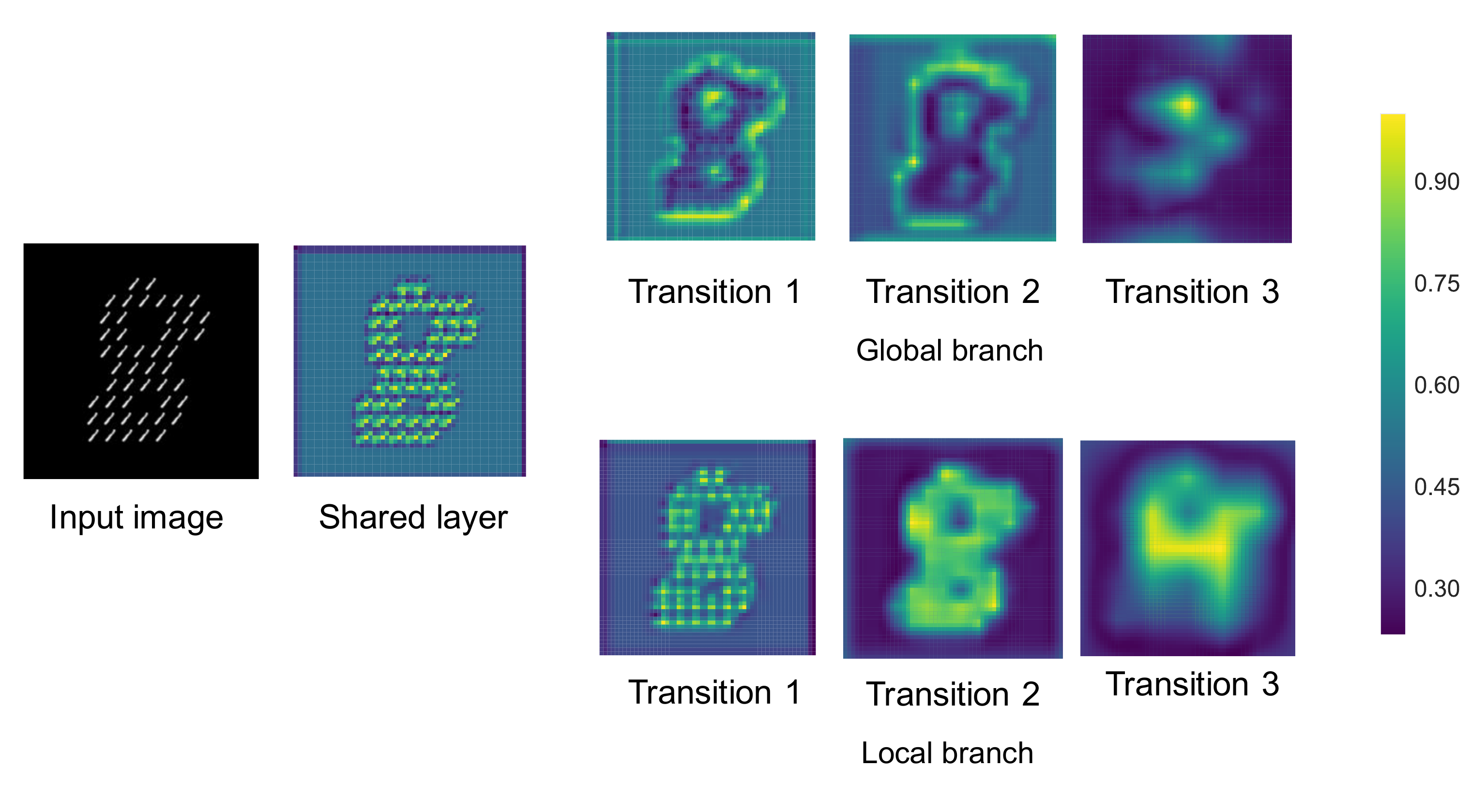}
	\small\caption{Feature visualization at shared and transition layers for Global and Local branches.}\label{fig:visualization}
\end{figure}
Investigating the responses of intermediate feature maps at different abstraction levels facilitates our understanding on nueral networks. Especially, the asymmetry of two branches leads to discrepant manipulatations on the given input. Here, we do feature visualization on different abstraction levels. Technically, we extract the output features from the first shared convolutional layer and transition layers of two branches. Then the absolute values of the feature maps are applied over by channel-wise average pooling and scaled to $[0, 1]$ for visualization. 

A case study on sb-MNIST dataset is shown in Fig.~\ref{fig:visualization}. From the visualization results, we can observe that the first shared features hold both global and local information. Then Global branch cares more about global context information and grasps the shape or style of big figure ``$8$'' very quickly at transition layer $1$ and $2$. Then the transformed features at transition layer $3$ of the Global branch represent the semantic information of the big figure. While the Local branch focuses more on the local details of the big figure and neglects the background instantly at transition layer $1$. And it keeps the shape of the big figure ``$8$'' almost unchanged until the last transition layer, which means the Local branch does not learn the pattern of the big figure. Instead, it transforms the local details, namely the features of small figures, into the final representation. 
\section{Conclusion}

Inspired by the recent study on the hemispheric specialization and coarse-to-fine perception, we proposed a novel left-right asymmetric layer skippable network for coarse-to-fine
object categorization. We leveraged a new design philosophy to make this network simultaneously classify coarse and fine-grained classes. In addition, we proposed the layer-skipping behavior of densely connected convolutional layers controlled by an auxiliary gating network. The experiments conducted on three datasets validate the performance, showing
the promising results of our proposed network.

\section*{Acknowledgement}

This work was supported by two projects from NSFC ($\#61622204$ and $\#61702108$) and two projects from STCSM ($\#16JC1420401$ and $\#16QA1400500$).

{\small
\bibliographystyle{ieee}
\bibliography{egbib}
}

\end{document}